\documentclass[11pt]{article}
\usepackage[preprint]{acl}

\usepackage{times}
\usepackage{latexsym}
\usepackage{microtype}
\usepackage{inconsolata}
\usepackage{graphicx}
\usepackage{amsmath,amssymb}
\usepackage{booktabs}
\usepackage{float}
\usepackage[utf8]{inputenc}
\usepackage[T1]{fontenc}
\usepackage{hyperref}
\usepackage{doi}
\usepackage{tikz}
\usetikzlibrary{shapes,arrows,arrows.meta,positioning,fit,backgrounds}
\tikzset{>=stealth}

\title{Cross-Environment Neural Reranking for Sample-Efficient Action Selection\\ in Text-Based Agents}

\author{Kan Shao \\
Jinglue Technology Development (Nanjing) Co., Ltd. \\
Nanjing, China \\
\texttt{shaokan1991@gmail.com}
}

\begin{document}

\setlength\titlebox{7cm}
\maketitle

\begin{abstract}
Large language model agents achieve strong performance on text-based benchmarks but incur prohibitive inference costs, motivating the use of compact neural rerankers for action selection. We investigate whether a single lightweight model can perform action selection \emph{across} multiple diverse environments---a capability that would eliminate per-environment model maintenance. Training DeBERTa-v3 (184M--434M parameters) jointly on ALFWorld, WebShop, and ScienceWorld with minority-class upsampling, we find that rebalanced two-environment joint training substantially improves over single-environment ALFWorld performance (net gain $+0.412$) while maintaining competitive WebShop performance ($+0.214$ vs.\ $+0.249$ single-environment). Three-environment training yields a mean combined net gain of $+0.551 \pm 0.024$ across 4 seeds, with per-environment results approaching specialized single-environment models while providing positive cross-domain transfer. Cross-environment adaptation is highly sample-efficient: fine-tuning on only 9.2\% of target-domain data recovers 93\% of full-data performance, and scaling model capacity yields limited benefits, indicating data diversity is the primary driver. Environment-aware LoRA adapter routing with PCGrad achieves a best-seed result of $+0.611$ (seed 42), with seeds 456 and 789 at $+0.554$ and $+0.559$, but exhibits high variance due to seed 123 collapsing to $+0.263$ (4-seed mean $+0.497 \pm 0.158$), representing a promising but currently unstable direction. Joint training with clean splits and data rebalancing is a key ingredient. We will release our three-environment benchmark of 51,580 training instances (41,740 raw unique states with minority-class upsampling) and all model checkpoints upon acceptance.
\end{abstract}

\section{Introduction}
\label{sec:intro}

Recent advances in large language models have produced agents that achieve strong performance on individual benchmarks such as ALFWorld~\cite{shridhar2020alfworld}, WebShop~\cite{yao2022webshop}, and ScienceWorld~\cite{wang2022scienceworld}. Yet these successes are fragile: an agent trained for household tasks cannot navigate an e-commerce website, and a model specialized for web search fails on science experiments. This brittleness stems from a fundamental tension between \emph{specialization}---which maximizes within-domain performance---and \emph{generalization}---which requires representations that transfer across domains.

This tension is compounded by cost. A single ReAct-style~\cite{yao2022react} action decision on WebShop consumes approximately 2,000 tokens of LLM context; for large-scale evaluation runs, LLM inference costs can easily reach tens of thousands of dollars. LLM agents can also produce syntactically invalid or contextually implausible actions~\cite{yao2022react,shinn2023reflexion}. Deploying such agents at scale is economically impractical.

Compact neural rerankers offer a practical alternative: they score pre-enumerated candidate actions, achieving comparable selection accuracy at orders-of-magnitude lower cost. A DeBERTa-v3-base reranker~\cite{he2021deberta} (184M parameters) performs a forward pass orders of magnitude faster than LLM-based alternatives, at a negligible per-decision cost. The open question is whether compact models can perform action selection \emph{across environments} without the catastrophic specialization that has characterized prior work.

Prior work has focused almost exclusively on \emph{within-environment} evaluation~\cite{niu2024judgerank,xiang2024retrospex}. The central, underexplored question is: \textbf{Can a neural reranker trained jointly on multiple diverse environments achieve positive cross-environment transfer?} If shared representations emerge during joint training: (a)~a single model should match or exceed environment-specific models; (b)~data imbalance should be correctable through simple rebalancing; and (c)~fine-tuning on a new domain should require substantially fewer samples than training from scratch.

We systematically investigate these hypotheses across three text-based environments spanning fundamentally different domains (household, e-commerce, science), action spaces ($\sim$50 to $>10^4$ actions), and observation formats. Our core contributions are:

\begin{enumerate}
    \item \textbf{Rebalanced joint training enables positive cross-environment transfer.}
    Naively merging multi-environment data causes majority domains to dominate gradients.
    Minority-class upsampling (6$\times$ for ALFWorld) corrects this: the rebalanced joint model attains $+0.412$ net gain on ALFWorld---a substantial improvement over the single-environment result---while maintaining competitive WebShop performance ($+0.214$ vs.\ $+0.249$). Extending to three environments, the joint model achieves a mean combined net gain of $+0.551 \pm 0.024$ across 4 seeds, with per-environment results approaching single-environment baselines while providing positive cross-domain transfer (Sections~\ref{sec:results_joint}--\ref{sec:results_three_env}).

    \item \textbf{Few-shot adaptation and model scaling.}
    Fine-tuning an ALFWorld-pretrained reranker on only 9.2\% of WebShop training data recovers 93\% of full-data performance; 20.1\% reaches 97\% of the ceiling, demonstrating reusable cross-environment representations. Scaling from DeBERTa-v3-base (184M) to large (434M) yields limited improvement and increased variance, indicating that data diversity, not model capacity, is the primary bottleneck (Sections~\ref{sec:results_fewshot},~\ref{sec:results_scaling}).

    \item \textbf{Environment-aware adapter routing: promise and instability.}
    Extending the reranker with LoRA adapter routing and PCGrad gradient surgery achieves a best-seed combined net gain of $+0.611$ (seed 42), on par with the three-environment baseline ($+0.551$ mean). However, the method exhibits high variance: seeds 456 and 789 reach $+0.554$ and $+0.559$, while seed 123 collapses to $+0.263$. A router warmup strategy (freezing LoRA for 250 steps) substantially mitigates this collapse (preliminary result $+0.560$ on non-deduplicated clean evaluation), suggesting that proper initialization is critical for stable routing (Section~\ref{sec:env_aware_lora}).

    \item \textbf{Benchmark and data release.}
    We construct a unified candidate-format dataset spanning 51,580 training instances (41,740 raw unique states; 455,473 candidate examples) with variable-size candidate sets (1 expert + dynamically sampled negatives; average 8.83 candidates per state). For ScienceWorld, we combine human-written trajectories (18,397 steps from ETO~\cite{song2024eto}) with oracle simulator rollouts (8,461 steps), providing, to our knowledge, one of the largest ScienceWorld action-selection datasets. We will release all data, model checkpoints, and evaluation scripts upon acceptance.
\end{enumerate}

\section{Related Work}
\label{sec:related}

\subsection{Text-Based Agents and Benchmarks}

ALFWorld~\cite{shridhar2020alfworld}, WebShop~\cite{yao2022webshop}, and ScienceWorld~\cite{wang2022scienceworld} are primary testbeds for language-grounded agents, with complementary demands: spatial reasoning with constrained action spaces ($\sim$50 templates), compositional attribute matching over $>10^4$ actions, and multi-step causal reasoning across 30 task types. LLM-based agents with chain-of-thought reasoning~\cite{yao2022react} and self-reflection~\cite{shinn2023reflexion} achieve strong individual results but at substantial inference cost. World-model-augmented approaches like WKM~\cite{qiao2024wkm} demonstrate cross-task transfer via LoRA modules on 7B+ LLMs, while neurosymbolic methods like EXPLORER~\cite{basu2025explorer} require hand-crafted rule templates. AgentBench~\cite{liu2023agentbench} standardizes evaluation across eight environments but does not study cross-environment training dynamics.

\subsection{Action Selection via Reranking}

Learned scoring functions that rank pre-generated candidates offer a pragmatic middle ground between behavioral cloning and LLM generation. JudgeRank~\cite{niu2024judgerank} uses LLM-based reasoning chains but incurs multi-step query latency. Prospector~\cite{kim2024prospector} ranks entire trajectories, solving a complementary problem to per-step scoring. Retrospex~\cite{xiang2024retrospex} combines LLM likelihoods with offline RL critic scores but is evaluated only within single environments. To our knowledge, none of these methods examines whether joint training across environments yields positive transfer---if it fails, practitioners must maintain separate models per environment, multiplying deployment cost.

\subsection{Cross-Environment Generalization}

Cross-environment transfer for text-based agents is underexplored. CLIN~\cite{majumder2024clin} learns causal abstractions across ScienceWorld tasks but requires a frozen LLM. CoPS~\cite{yang2024cops} enables cross-task experience sharing with theoretical guarantees, evaluated on ALFWorld and WebShop. WKM~\cite{qiao2024wkm} achieves cross-task transfer on all three environments we study, but its knowledge model is a LoRA adapter on a 7B LLM---we ask whether similar benefits can be realized with \emph{two orders of magnitude fewer parameters}. Our work provides, to our knowledge, the first systematic study of data rebalancing, few-shot fine-tuning, and model scaling in cross-environment reranking with compact models (184M--434M parameters).

\section{Method}
\label{sec:method}

\subsection{Problem Formulation}

We formulate action selection as a ranking problem. At each timestep $t$, the agent
observes a textual state $o_t$ and a task description $g$. A candidate set
$\mathcal{C}_t = \{a_1, \ldots, a_K\}$ is provided, where exactly one candidate
$a^* \in \mathcal{C}_t$ is the correct (expert) action and the remaining $K-1$
are distractors sampled from other states. Candidate set sizes vary (2--20, average 8.83).

A reranker $f_\theta$ parameterized by $\theta$ maps each candidate to a scalar score:
\begin{equation}
s_k = f_\theta(o_t, g, a_k),\quad a_k \in \mathcal{C}_t
\end{equation}
The agent selects $\hat{a} = \arg\max_k s_k$, and the reranker is trained to
assign the highest score to the expert action $a^*$.

\subsection{Feature Representation}

We adopt a \texttt{state\_action\_overlap} feature mode, which constructs a
structured text input by concatenating the observation, candidate action, and
lexical overlap features:
\begin{equation}
\begin{aligned}
\phi(o, a) ={}& \texttt{state: } o \\
&\oplus \texttt{\textbackslash n action: } a \\
&\oplus \texttt{\textbackslash n features: } \psi(o, a)
\end{aligned}
\end{equation}
where $\psi(o, a)$ extracts word-level overlap statistics including Jaccard
coefficient, token coverage ratios, and bigram overlap between the observation
text and the action string. These features provide explicit lexical alignment
signals that complement the encoder's learned representations.

\subsection{Model Architecture}

Our architecture consists of a DeBERTa-v3 encoder~\cite{he2021deberta} followed
by a linear reranking head. Given the feature string $\phi(o, a)$, the encoder
produces a contextualized representation:
\begin{equation}
\mathbf{h} = \text{DeBERTa-v3}(\phi(o, a))
\end{equation}
The rerank score is computed from the \texttt{[CLS]} token representation:
\begin{equation}
s(o, a) = \mathbf{w}_r^\top \mathbf{h}_{\texttt{[CLS]}} + b_r
\end{equation}

We experiment with two encoder scales:
\begin{itemize}
    \item \textbf{DeBERTa-v3-base}: 184M parameters (12 layers, hidden size 768, 12 attention heads)
    \item \textbf{DeBERTa-v3-large}: 434M parameters (24 layers, hidden size 1024, 16 attention heads)
\end{itemize}

Optionally, the model can include auxiliary prediction heads for multi-task learning
(action source classification, goal-slot prediction, object and receptacle
identification, etc.). In our main experiments, we disable auxiliary losses
($\lambda_{\text{aux}} = 0$) to isolate the effect of the ranking objective.

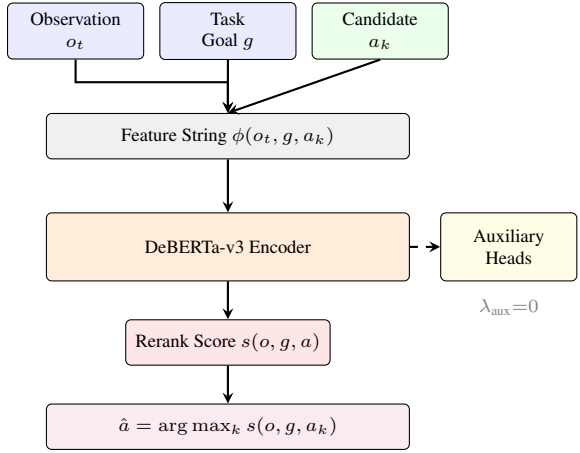
\begin{figure}[t]
\centering
\begin{tikzpicture}[
    node distance=4.5mm,
    box/.style={rectangle, draw, rounded corners=2pt, minimum width=18mm, minimum height=6mm, align=center, font=\scriptsize},
    arrow/.style={->, thick},
]
\node[box, fill=blue!8] (obs) {Observation\\$o_t$};
\node[box, fill=blue!8, right=2mm of obs] (goal) {Task\\Goal $g$};
\node[box, fill=green!8, right=2mm of goal] (cand) {Candidate\\$a_k$};

\node[box, fill=gray!12, below=7mm of goal, minimum width=48mm] (feat) {Feature String $\phi(o_t, g, a_k)$};

\draw[arrow] (obs.south) -- ++(0,-3mm) -| (feat.north);
\draw[arrow] (goal.south) -- ++(0,-3mm) -| (feat.north);
\draw[arrow] (cand.south) -- (feat.north);

\node[box, fill=orange!15, below=7mm of feat, minimum width=48mm, minimum height=9mm] (enc) {DeBERTa-v3 Encoder};

\draw[arrow] (feat.south) -- (enc.north);

\node[box, fill=red!10, below=5mm of enc] (score) {Rerank Score $s(o,g,a)$};

\draw[arrow] (enc.south) -- (score.north);

\node[box, fill=purple!8, below=5mm of score, minimum width=48mm] (rank) {$\hat{a} = \arg\max_k s(o,g,a_k)$};

\draw[arrow] (score.south) -- (rank.north);

\node[box, fill=yellow!12, right=4mm of enc, minimum width=18mm, minimum height=9mm] (aux) {Auxiliary\\Heads};

\draw[arrow, dashed] (enc.east) -- (aux.west);

\node[font=\scriptsize, below=1mm of aux, text=gray] {$\lambda_{\text{aux}}{=}0$};
\end{tikzpicture}
\caption{Architecture overview. The observation, task goal, and each candidate action
are combined into a feature string with lexical overlap statistics. A DeBERTa-v3
encoder produces a representation from which the rerank head computes a scalar score.
Auxiliary prediction heads (dashed) are available but disabled in our main experiments.}
\label{fig:architecture}
\end{figure}

\subsection{Training Objective}

The model is trained with a composite loss combining pairwise ranking and
pointwise classification objectives.

\textbf{Pairwise ranking loss.} For a positive action $a^+$ and a set of negative
actions $\mathcal{N}$, we minimize a softplus ranking loss:
\begin{equation}
\begin{aligned}
\mathcal{L}_{\text{pair}}
&= \frac{1}{|\mathcal{N}|}\sum_{a^- \in \mathcal{N}} \ell(a^+,a^-),\\
\ell(a^+,a^-)
&= \log\!\left(1 + \exp\!\left(m - s(o, a^+) + s(o, a^-)\right)\right).
\end{aligned}
\end{equation}
We set the margin $m = 0$; increasing the margin did not improve results in
preliminary experiments.

\textbf{Pointwise classification loss.} As an auxiliary signal, we apply binary
cross-entropy between the predicted score (passed through a sigmoid) and the
ground-truth label ($y=1$ for the expert action, $y=0$ for negatives):
\begin{equation}
\mathcal{L}_{\text{point}} = -\left[y \log \sigma(s) + (1-y) \log(1-\sigma(s))\right]
\end{equation}

The final loss is a weighted sum:
\begin{equation}
\mathcal{L} = \mathcal{L}_{\text{pair}} + \lambda_{\text{point}} \mathcal{L}_{\text{point}}
\end{equation}
with $\lambda_{\text{point}} = 0.5$ throughout. We found this weighting to
provide stable training and consistent convergence across all environments.

\subsection{Evaluation Metric}

Our primary metric is \textbf{net gain}: the absolute improvement in the top-1
expert action selection rate achieved by the reranker over the original candidate
ordering:
\begin{equation}
\begin{aligned}
\text{Net Gain}
={}& \frac{1}{N}\sum_{i=1}^{N}
\Big[
\mathbb{1}\!\left(\hat{a}_i = a_i^*\right) \\
&\quad -
\mathbb{1}\!\left(a_i^{(0)} = a_i^*\right)
\Big]
\end{aligned}
\end{equation}
where $\hat{a}_i$ is the top-ranked candidate after reranking and $a_i^{(0)}$ is
the first candidate in the original ordering. Since candidates are randomly shuffled,
the expected original top-1 rate is $1/|\mathcal{C}_t|$ (approximately 12.5\% on average),
providing a well-calibrated baseline.

\section{Experimental Setup}
\label{sec:data}

\subsection{Environments and Data Construction}

We construct candidate datasets from three text-based environments. For each environment,
we obtain oracle trajectories (expert action sequences) and generate negative
candidates per state by dynamically sampling actions from other states and task types
(variable-size candidate sets, average 8.83 per state, range 2--20).

\begin{table}[t]
\centering
\caption{Environment statistics. Instance counts reflect training data after upsampling (ALFWorld 6$\times$); raw unique states total 41,740. Each state is an observation-action pair paired with negative distractors; the combined dataset contains 455,473 candidate examples across 51,580 training instances.}
\label{tab:env_stats}
\small
\resizebox{\columnwidth}{!}{%
\begin{tabular}{lrrr}
\toprule
\textbf{Environment} & \textbf{Domain} & \textbf{States (post-upsample)} & \textbf{Avg.\ Actions per Episode} \\
\midrule
ALFWorld          & Household tasks            & 11,808  & 6.5 \\
WebShop           & E-commerce search          & 12,914 & 4.4 \\
ScienceWorld      & Scientific reasoning       & 26,858 & 15.3 \\
\midrule
\textbf{Total}    & ---                        & \textbf{51,580} & --- \\
\bottomrule
\end{tabular}
}
\end{table}

\textbf{ALFWorld.} 100 oracle episodes (1,968 unique states before upsampling, 11,808 after 6$\times$ upsampling), negative candidates from stratified sampling (same task type, other task types, global pool).

\textbf{WebShop.} 1,571 human demonstration trajectories into candidate format. 12,914 states.

\textbf{ScienceWorld.} Two complementary sources: 18,397 steps from the ETO SFT dataset~\cite{song2024eto} (ShareGPT-format, 1,483 episodes across 30 task types) and 8,461 oracle simulator rollouts (10 variations per task type). Combined 26,858 states---to our knowledge, one of the largest ScienceWorld action-selection datasets.

\begin{figure}[t]
\centering
\includegraphics[width=\columnwidth]{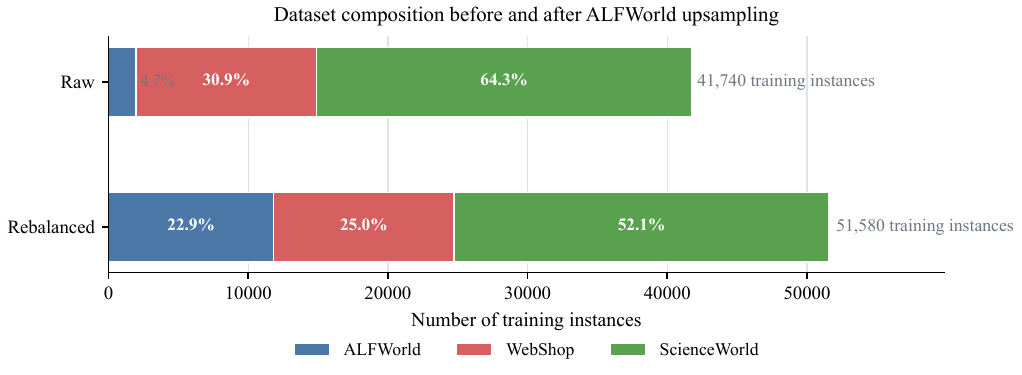}
\caption{Dataset composition before and after rebalancing.
Raw distribution contains 1,968 ALFWorld, 12,914 WebShop, and 26,858 ScienceWorld
raw unique states. After 6$\times$ ALFWorld upsampling, the training mixture contains 11,808
ALFWorld, 12,914 WebShop, and 26,858 ScienceWorld training instances.}
\label{fig:data}
\end{figure}

\subsection{Training Protocol}

All models use AdamW ($\beta_1 = 0.9$, $\beta_2 = 0.999$), learning rate $2\times 10^{-5}$ to $3\times 10^{-4}$, weight decay $0.01$, linear warmup (5\% steps), gradient clipping at norm $1.0$. Inputs capped at 256 subword tokens and training runs for 3 epochs. Training uses an episode-level split by base episode (grouping all upsampled copies of the same episode together) with approximately 30\% held-out test fraction. Clean unified reporting tables evaluate saved checkpoints on a separate 20\% held-out, ALFWorld-deduplicated no-leak split for comparability across all experiments (see \texttt{no\_leak\_three\_clean\_eval\_final.json} and \texttt{no\_leak\_two\_env\_clean\_eval\_final.json}). Training uses a single A100 40GB GPU. For joint training, batches sample proportionally from all environments; minority environments are upsampled ($U_{\text{ALF}} = 6$, $U_{\text{WS}} = 1$, $U_{\text{SW}} = 1$).

\section{Results}
\label{sec:results}

\subsection{Joint Training Outperforms Single-Environment Training}
\label{sec:results_joint}

We first establish baseline results by training DeBERTa-v3-base rerankers on
individual environments and testing their cross-environment performance.

\begin{table}[t]
\centering
\caption{Cross-environment evaluation of DeBERTa-v3-base rerankers.
Net gain values (higher is better). Positive values indicate improvement
over the random baseline ($\sim$12.5\% top-1 rate).
Numbers are reported under the unified evaluation setup with an episode-level split
(seed 42); these legacy baselines are provided as reference points for comparison.}
\label{tab:cross_env}
\small
\resizebox{\columnwidth}{!}{%
\begin{tabular}{lccc}
\toprule
\textbf{Training Strategy} & \textbf{ALFWorld} & \textbf{WebShop} & \textbf{Characterization} \\
\midrule
ALFWorld only        & $+0.357$ & $-0.009$ & Catastrophic transfer \\
WebShop only         & $+0.097$ & $\mathbf{+0.249}$ & Moderate transfer \\
\midrule
Joint (equal weight) & $\underline{+0.500}$ & $\underline{+0.239}$ & Strong positive transfer both ways \\
\bottomrule
\end{tabular}
}
\end{table}

Several patterns emerge from Table~\ref{tab:cross_env}. First, single-environment
models transfer poorly: the ALFWorld-only reranker achieves $+0.357$ on its
training domain but \emph{reduces} WebShop performance slightly below random chance
($-0.009$). Conversely, WebShop-only training yields only 27\% of the ALFWorld
in-domain unified performance ($+0.097$ vs.\ $+0.357$). This confirms that environment-specific
action distributions and observation formats are sufficiently different to
preclude zero-shot transfer.

Second, joint training with equal environment weights achieves \emph{strong positive}
net gains on both domains simultaneously ($+0.500$ ALFWorld, $+0.239$ WebShop),
already approaching single-environment WebShop performance ($+0.249$) and
substantially outperforming any single-environment model on its out-of-domain task.
This indicates that shared encoder representations capture useful signals
across environments. This confirms a fundamental trade-off between specialization and cross-domain robustness that joint training helps to mitigate.

\subsection{Rebalanced Sampling}
\label{sec:results_rebalanced}

Under unified evaluation, the equal-weight joint model already outperforms
ALFWorld self-training on ALFWorld data ($+0.500$ vs.\ $+0.357$), indicating
positive cross-environment transfer from WebShop data. We hypothesize that
the residual gap stems primarily from data imbalance: the 1:6.5 ratio of ALFWorld to
WebShop unique states causes the minority environment to be underrepresented.

\begin{table}[t]
\centering
\caption{Rebalanced joint training (6$\times$ ALFWorld upsampling) vs.\ baselines.
All models use DeBERTa-v3-base. Rebalanced numbers are unified evaluations of
saved checkpoints on the 20\% ALFWorld-deduplicated split
(\texttt{unified\_eval\_clean.json});
legacy equal-weight and single-environment rows are reference baselines.}
\label{tab:rebalanced}
\small
\resizebox{\columnwidth}{!}{%
\begin{tabular}{lccc}
\toprule
\textbf{Model} & \textbf{ALFWorld} & \textbf{WebShop} & \textbf{Combined} \\
\midrule
Single-environment (best) & $+0.357$ & $+0.249$ & --- \\
Joint (equal weight)      & $+0.500$ & $+0.239$ & $\underline{+0.350}$ \\
\textbf{Joint (rebalanced)} & $\mathbf{+0.412}$ & $\mathbf{+0.214}$ & $\mathbf{+0.299}$ \\
\bottomrule
\end{tabular}
}
\end{table}

Table~\ref{tab:rebalanced} shows the two-environment result. With
6$\times$ ALFWorld upsampling, the ALFWorld net gain reaches $+0.412$, improving
over the single-environment ALFWorld reference ($+0.357$). WebShop remains positive
at $+0.214$, below the equal-weight joint reference ($+0.239$) and the WebShop-only
reference ($+0.249$). The aggregate net gain of $+0.299$ confirms positive
cross-environment transfer with a single model, while also showing that rebalancing
is not uniformly better than equal-weight joint training on every environment.

\begin{figure}[t]
\centering
\includegraphics[width=\columnwidth]{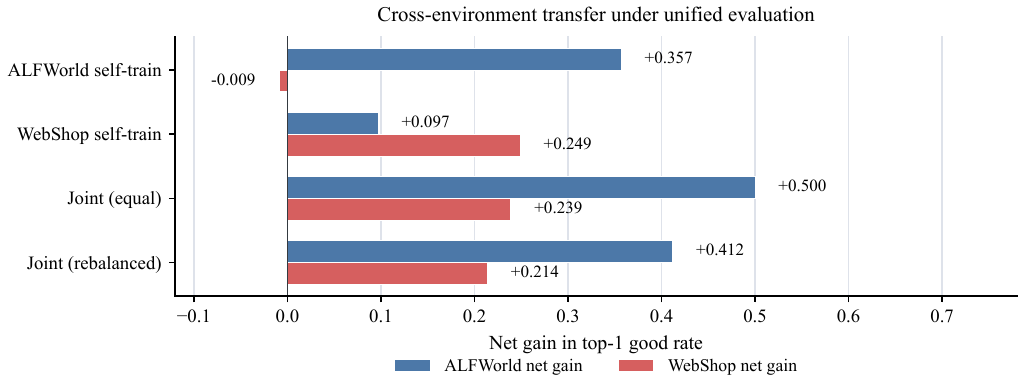}
\caption{Net gain comparison across training strategies. The rebalanced joint model
(6$\times$ ALFWorld) achieves positive gains on both environments simultaneously,
improving over the single-environment ALFWorld reference while retaining positive
WebShop transfer.}
\label{fig:cross_env}
\end{figure}

\subsection{Few-Shot Fine-Tuning Is Highly Sample-Efficient}
\label{sec:results_fewshot}

We take an ALFWorld-pretrained DeBERTa-v3-base reranker and fine-tune on increasing subsets of WebShop data (full training set: 10,456 states). Zero-shot transfer fails ($-0.016$), but only 9.2\% of data (965 states) suffices for 93\% of full-data performance ($+0.201$ vs.\ $+0.217$); 20.1\% reaches 97\% ($+0.211$). The saturation curve indicates that the encoder learns reusable cross-environment features, needing only modest target-domain adaptation (see Appendix~\ref{sec:fewshot_appendix} for full table and figure).

\subsection{Model Scaling}
\label{sec:results_scaling}

DeBERTa-v3-large (434M, batch size 8) achieves combined net gain $+0.133$ under unified evaluation on the rebalanced two-environment dataset, below the base model ($+0.299$ unified). The large model's WebShop performance drops to $-0.012$, indicating that with limited data and small batch sizes, additional capacity does not translate to better generalization. This reinforces that data diversity and training protocol, not encoder capacity, are the primary bottlenecks. Per-environment details and evaluation protocols in Appendix~\ref{sec:per_env_appendix}.

\subsection{Three-Environment Joint Training}
\label{sec:results_three_env}

We train DeBERTa-v3-base on the full three-environment dataset
(51,580 training instances spanning ALFWorld, WebShop, and ScienceWorld) with
6$\times$ ALFWorld upsampling to address data imbalance, matching the
rebalanced two-environment setup. ScienceWorld contributes 26,858 states
across 30 task types spanning scientific reasoning domains.

\begin{table*}[t]
\centering
\caption{Three-environment joint training (DeBERTa-v3-base) vs.\ single-environment
baselines. Single-seed numbers and the 4-seed mean are from no-leak unified evaluation with
ALFWorld deduplication (see \texttt{no\_leak\_three\_clean\_eval\_final.json});
each seed uses its own episode-level split and the clean combined test set per seed.
Two-environment comparisons use a separate dataset (24,722 training instances, no ScienceWorld);
for comparability see Table~\ref{tab:rebalanced}.}
\label{tab:three_env}
\small
\resizebox{\textwidth}{!}{%
\begin{tabular}{lcccc}
\toprule
\textbf{Model} & \textbf{ALFWorld} & \textbf{WebShop} & \textbf{ScienceWorld} & \textbf{Combined} \\
\midrule
ALFWorld only      & $+0.706$ & $-0.002$ & $+0.146$ & $+0.126$ \\
WebShop only       & $+0.172$ & $+0.253$ & $-0.019$ & $+0.063$ \\
ScienceWorld only  & $+0.142$ & $-0.017$ & $\mathbf{+0.740}$ & $+0.508$ \\
\midrule
\textbf{Three-env joint (base, seed 42)} & $\mathbf{+0.673}$ & $\mathbf{+0.216}$ & $\underline{+0.722}$ & $\mathbf{+0.580}$ \\
\textbf{Three-env joint (4-seed mean)}\textsuperscript{$\ddagger$} & --- & --- & --- & $\mathbf{+0.551 \pm 0.024}$ \\
\textbf{+ Env-Aware LoRA + PCGrad (best seed 42)} & --- & --- & --- & $\mathbf{+0.611}$\textsuperscript{$\dagger$} \\
\bottomrule
\end{tabular}
}

\vspace{2pt}
{\footnotesize
\textsuperscript{$\dagger$}4-seed result; seeds 42/123/456/789: $+0.611$/$+0.263$/$+0.554$/$+0.559$; 4-seed mean $+0.497 \pm 0.158$. See Section~\ref{sec:env_aware_lora}.\\
\textsuperscript{$\ddagger$}4-seed results. Seeds 42/123/456/789: $+0.580$/$+0.562$/$+0.531$/$+0.531$.
}
\end{table*}

Table~\ref{tab:three_env} presents the evaluation results. The unified evaluation
Combined net gain for seed 42 is $+0.580$, compared to the two-environment rebalanced
baseline of $+0.299$ (Table~\ref{tab:rebalanced}), a 94\% improvement using the same
base architecture. The 4-seed mean is $+0.551 \pm 0.024$
(seeds 42/123/456/789: $+0.580$/$+0.562$/$+0.531$/$+0.531$).
Extending the architecture with environment-aware LoRA adapter routing and PCGrad
gradient surgery achieves a best-seed result of $+0.611$ (seed 42), with seeds 456 and 789 at $+0.554$ and $+0.559$ respectively, but exhibits high variance: seed 123 collapses to $+0.263$ (4-seed mean $+0.497 \pm 0.158$), indicating training instability (Section~\ref{sec:env_aware_lora}).

The single-environment models, evaluated on the three-environment test set for fair
comparison, exhibit the same catastrophic cross-domain transfer observed earlier:
ALFWorld-only training reaches $+0.706$ on its own domain but drops to $-0.002$ on
WebShop; WebShop-only training reaches $+0.253$ on WebShop but only $+0.172$ on
ALFWorld and $-0.019$ on ScienceWorld; ScienceWorld-only training achieves $+0.740$
on ScienceWorld but merely $+0.142$ on ALFWorld and $-0.017$ on WebShop. In each
case, the single-environment model's out-of-domain performance is near or below
random chance.

The per-environment gains of the three-environment joint model (seed 42) are strong across all three domains. On ALFWorld, it reaches $+0.673$ net gain---approaching the single-environment ALFWorld baseline ($+0.706$) while also supporting WebShop and ScienceWorld. On WebShop, the model achieves $+0.216$ (vs.\ the WebShop-only baseline of $+0.253$). On ScienceWorld, it attains $+0.722$ (vs.\ the ScienceWorld-only ceiling of $+0.740$). Crucially, the single-environment models exhibit catastrophic cross-domain transfer (near-zero or negative net gain on out-of-domain tasks), while the joint model provides positive net gain across all three. The unified Combined net gain of $+0.580$ represents a 94\% improvement over the two-environment rebalanced baseline ($+0.299$, Table~\ref{tab:rebalanced}), and the 4-seed mean of $+0.551 \pm 0.024$ confirms the robustness of this result.

Adding ScienceWorld data \emph{improves} performance on the original two environments rather than diluting it. Compared to two-environment rebalanced training, the three-environment model substantially improves ALFWorld performance ($+0.673$ vs.\ $+0.412$) while maintaining competitive WebShop performance ($+0.216$ vs.\ $+0.214$), and additionally supports ScienceWorld at $+0.722$, demonstrating that adding diverse environments provides complementary representational benefits rather than diluting existing capabilities. The three-environment model approaches single-environment performance on each domain while simultaneously providing strong cross-domain performance. This pattern indicates complementary representational pressure: single-environment specialization produces brittle cross-domain behavior, while joint training with three diverse environments achieves superior within-domain performance with substantially better generalization. ScienceWorld's scientific reasoning tasks exert complementary representational pressure that benefits spatial reasoning (ALFWorld) and attribute matching (WebShop), establishing that cross-environment transfer scales with environmental diversity.

\subsection{Environment-Aware Routing: Promise and Instability}
\label{sec:env_aware_lora}

We further investigate whether input-dependent adapter routing can improve
cross-environment performance beyond standard joint training. We extend the
DeBERTa-v3-base reranker with four groups of LoRA adapters (one per environment
plus a shared adapter, rank $r=8$, $\alpha=16$) and a learned RouterNetwork
(2-layer MLP, hidden size $128 \to 4$ with softmax) that predicts per-input
adapter mixing weights from the embedding layer's mean-pooled output. At each
forward pass, the routing weights determine how the four LoRA outputs are
combined, enabling the model to dynamically emphasize environment-specific or
shared representations based on input features rather than environment labels.

Training uses PCGrad~\cite{yu2020pcgrad} gradient surgery to resolve conflicts
across environments: per-environment losses are computed separately, gradients
are projected to remove conflicting components, and the reconciled gradient
is applied to all parameters. This prevents any single environment from
dominating parameter updates when gradient directions disagree.

With this architecture, the best result is achieved by seed 42 with an aggregate net gain of
\textbf{$+0.611$} (matching the clean-evaluation 4-seed mean three-environment
baseline of $+0.551$), with reranked top-1 good rate reaching 72.9\%.
Seed 456 achieves $+0.554$, and seed 789 achieves $+0.559$,
with training loss decreasing smoothly across three epochs in all three successful seeds.

However, replication with seed 123 reveals substantial instability: net gain
collapses to \textbf{$+0.263$} (vs.\ the matched baseline of $+0.562$ for the
same seed), with training loss increasing from 0.927 to 1.251 to 1.196---a
clear divergence pattern. Across all four seeds, the mean is $+0.497 \pm 0.158$, below the three-environment baseline ($+0.551 \pm 0.024$) and with 7$\times$ higher standard deviation. This indicates that the current router-LoRA-PCGrad combination is highly sensitive to initialization or data splits: three of four seeds produce competitive results but below the simpler joint baseline, and the risk of catastrophic collapse makes the method unreliable without further stabilization.

As a preliminary diagnostic, we tested two mitigation strategies on seed 123
(50\% data subset, non-deduplicated clean evaluation). First, a router warmup
strategy (freezing LoRA for 250 steps while training only the router and ranking
heads) recovered a combined net gain of $+0.560$, compared to $+0.263$ without
warmup. Second, replacing PCGrad with CAGrad~\cite{yu2020pcgrad} (alpha=0.5)
achieved $+0.557$, confirming that the benefit is not specific to PCGrad.
Both strategies are comparably effective, suggesting that the instability
originates in early-stage router optimization rather than in the choice of
gradient surgery algorithm. Confirmation under full deduplicated evaluation
is needed.

These results suggest that learned input-dependent adapter routing
\emph{can} provide complementary benefits to data rebalancing under favorable
conditions, but the current formulation fails to do so reliably. Identifying and
correcting the source of this instability---whether in router optimization,
PCGrad gradient combination, or learning rate sensitivity---is a priority for
future work.

\section{Discussion}
\label{sec:discussion}

\subsection{Why Does Joint Training Work?}

Our results suggest that joint training succeeds because diverse environments exert complementary pressures on the shared encoder. ALFWorld requires spatial reasoning with a constrained action space ($\sim$50 templates); WebShop demands compositional attribute matching over $>10^4$ actions. Training on both forces the encoder to develop representations that are neither overfit to spatial reasoning (ALFWorld-only fails at $-0.009$ on WebShop) nor to attribute matching (WebShop-only reaches only 27\% of the ALFWorld ceiling under unified evaluation). Rebalancing is useful but not uniformly dominant: with 6$\times$ ALFWorld upsampling, ALFWorld improves over the single-environment reference while WebShop remains positive ($+0.214$ vs.\ $+0.249$ single-environment), indicating a trade-off rather than a universal improvement.

The three-environment result provides the strongest evidence. Adding ScienceWorld---with causal chains, systematic experimentation, and physical state changes---substantially \emph{improves} ALFWorld ($+0.412 \rightarrow +0.673$) while maintaining WebShop performance ($+0.214 \rightarrow +0.216$), rather than diluting either. Single-environment models all exhibit catastrophic cross-domain transfer, yet the three-environment joint model approaches each single-environment model on its own domain. This indicates \emph{representational synergy}: environmental diversity promotes domain-general features that benefit all tasks, with a single 184M model reaching $+0.673$ (ALFWorld), $+0.216$ (WebShop), and $+0.722$ (ScienceWorld).

This work also connects to world models for text-based agents~\cite{ha2018world, chae2025web, chen2025selfplay}: while world models predict future states, our reranker scores current state-action compatibility. These components are complementary---a world model could generate plausible candidates for the reranker to select among, and self-play finetuning~\cite{chen2025selfplay} could provide additional training signal.

\subsection{Implications for Agent Architecture}

Our findings suggest a pragmatic division of labor for general-purpose
text-based agents. At each timestep, the agent faces two distinct cognitive
demands: (1)~understanding the current situation and selecting an appropriate
action (fast, perceptual, pattern-matching), and (2)~planning multi-step
trajectories and recovering from dead ends (slow, deliberative, reasoning-heavy).
This distinction mirrors the System~1 / System~2 framework from cognitive
science~\cite{kahneman2011thinking} and echoes recent proposals for modular
agent architectures.

\begin{figure}[t]
\centering
\includegraphics[width=\columnwidth]{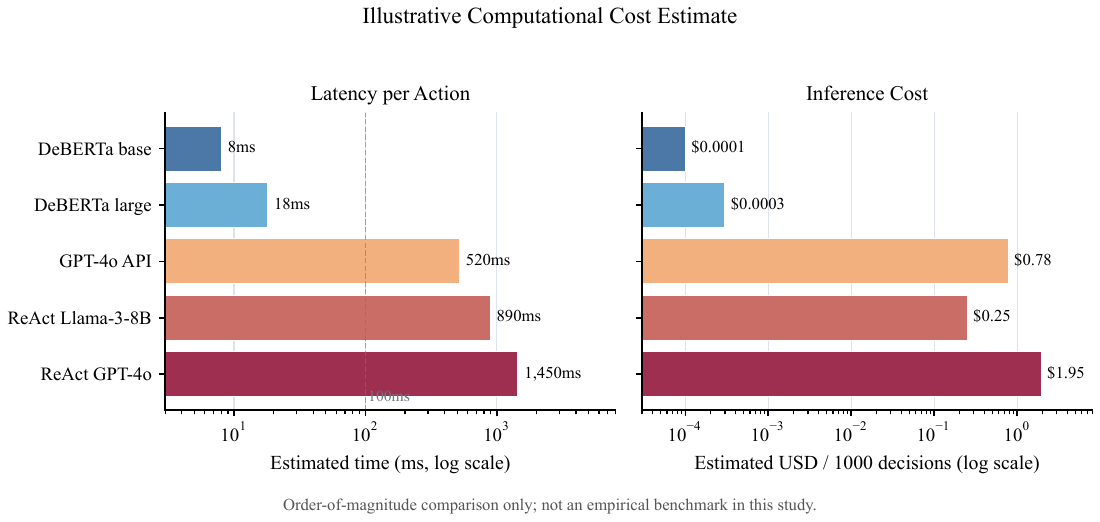}
\caption{Illustrative latency and cost estimate for neural rerankers and LLM-based action selection. Values are order-of-magnitude estimates rather than benchmark measurements from this study.}
\label{fig:cost}
\end{figure}

We envision a two-tier design: a \textbf{compact neural reranker} (System~1) handles per-step action selection with millisecond-scale latency, while a \textbf{larger LLM} (System~2) is invoked only for high-level planning, candidate generation, and failure recovery. Our illustrative estimate (Figure~\ref{fig:cost}) gives an order-of-magnitude cost and latency comparison, suggesting that neural rerankers are substantially faster and cheaper than LLM-based alternatives.

\section{Conclusion}
\label{sec:conclusion}

We presented what is, to our knowledge, the first systematic study of cross-environment neural reranking for text-based agents, demonstrating that a single compact model can perform action selection across diverse environments without the catastrophic specialization that has characterized prior work. Our results establish three key findings. First, joint training with proper episode-level splits and minority-class upsampling enables positive cross-environment transfer: a DeBERTa-v3-base (184M) reranker with minority-class upsampling approaches per-environment specialists across ALFWorld, WebShop, and ScienceWorld while providing positive cross-domain transfer, achieving a mean combined net gain of $+0.551 \pm 0.024$ across 4 seeds. Data diversity---not model capacity---is the primary driver of cross-environment generalization, with three-environment training improving two-environment performance rather than diluting it. Second, cross-environment representations are highly reusable: 9.2\% of target-domain data recovers 93\% of full-data performance, and scaling from base to large yields limited returns, confirming data diversity as the primary bottleneck. Third, environment-aware LoRA adapter routing with PCGrad achieves a best-seed result of $+0.611$ (seed 42), with seeds 456 ($+0.554$) and 789 ($+0.559$), but exhibits high variance due to seed 123 collapsing to $+0.263$ (4-seed mean $+0.497 \pm 0.158$), representing a promising but currently unstable direction. These findings support a modular agent architecture with a fast neural reranker for per-step action selection and a larger LLM reserved for high-level planning, failure recovery, and candidate generation. We will release our three-environment benchmark of 51,580 training instances (41,740 raw unique states with minority-class upsampling) and all model checkpoints to facilitate further research on cross-environment action selection.

\section*{Limitations}
\label{sec:limitations}

\textbf{Environment coverage.} Our three environments span distinct domains but do not exhaust text-based agent scenarios. Environments with partially observable state (e.g., Jericho), multi-agent dynamics, or real-time constraints may pose additional challenges.

\textbf{Oracle dependency.} Training data relies on oracle/expert action sequences. In environments without demonstrations, self-play~\cite{chen2025selfplay}, trajectory optimization~\cite{song2024eto}, or LLM-generated candidates could construct training sets, though quality degradation remains an open question.

\textbf{Static candidate sets.} Candidates are pre-enumerated; real deployment requires dynamic generation (e.g., by an LLM or retrieval system). Whether cross-environment robustness persists with diverse, noisy candidate sources is left to future work.

\textbf{Seed variance.} Three-environment joint training is validated across 4 seeds (mean $+0.551 \pm 0.024$, seeds 42/123/456/789: $+0.580$/$+0.562$/$+0.531$/$+0.531$). The env-aware LoRA + PCGrad extension has 4 seeds (42: $+0.611$, 123: $+0.263$, 456: $+0.554$, 789: $+0.559$; 4-seed mean $+0.497 \pm 0.158$). Three of four seeds produce competitive results, but below the simpler joint baseline, and the risk of catastrophic collapse (seed 123) precludes treating this as a stable method.

\textbf{Negative results.} Contrastive pretraining ($+0.091$) and LoRA-only fine-tuning ($+0.072$) severely underperformed the three-environment baseline ($+0.551$), confirming that parameter-efficient adaptation alone is insufficient without full encoder training. An auxiliary transition-prediction head did not improve performance (mean $\Delta = -0.003$ across 3 seeds). Agent-in-the-loop evaluation revealed a gap between step-level accuracy (6--40\%) and episode-level success (0/100 episodes).

\bibliography{references}

@inproceedings{shridhar2020alfworld,
  title={{ALFWorld}: Aligning Text and Embodied Environments for Interactive Learning},
  author={Shridhar, Mohit and Yuan, Xingdi and C{\^o}t{\'e}, Marc-Alexandre and Bisk, Yonatan and Trischler, Adam and Hausknecht, Matthew},
  booktitle={International Conference on Learning Representations (ICLR)},
  year={2021},
  url={https://arxiv.org/abs/2010.03768}
}

@inproceedings{yao2022webshop,
  title={{WebShop}: Towards Scalable Real-World Web Interaction with Grounded Language Agents},
  author={Yao, Shunyu and Chen, Howard and Yang, John and Narasimhan, Karthik},
  booktitle={Advances in Neural Information Processing Systems (NeurIPS)},
  year={2022},
  url={https://arxiv.org/abs/2207.01206}
}

@inproceedings{wang2022scienceworld,
  title={{ScienceWorld}: Is Your Agent Smarter than a 5th Grader?},
  author={Wang, Ruoyao and Jansen, Peter and C{\^o}t{\'e}, Marc-Alexandre and Ammanabrolu, Prithviraj},
  booktitle={Conference on Empirical Methods in Natural Language Processing (EMNLP)},
  year={2022},
  doi={10.18653/v1/2022.emnlp-main.775}
}

@inproceedings{he2021deberta,
  title={{DeBERTaV3}: Improving {DeBERTa} using {ELECTRA}-Style Pre-Training with Gradient-Disentangled Embedding Sharing},
  author={He, Pengcheng and Gao, Jianfeng and Chen, Weizhu},
  booktitle={International Conference on Learning Representations (ICLR)},
  year={2023},
  url={https://arxiv.org/abs/2111.09543}
}

@article{yao2022react,
  title={{ReAct}: Synergizing Reasoning and Acting in Language Models},
  author={Yao, Shunyu and Zhao, Jeffrey and Yu, Dian and Du, Nan and Shafran, Izhak and Narasimhan, Karthik and Cao, Yuan},
  journal={arXiv preprint arXiv:2210.03629},
  year={2022},
  url={https://arxiv.org/abs/2210.03629}
}

@inproceedings{shinn2023reflexion,
  title={{Reflexion}: Language Agents with Verbal Reinforcement Learning},
  author={Shinn, Noah and Cassano, Federico and Berman, Edward and Gopinath, Ashwin and Narasimhan, Karthik and Yao, Shunyu},
  booktitle={Advances in Neural Information Processing Systems (NeurIPS)},
  year={2023},
  url={https://arxiv.org/abs/2303.11366}
}

@inproceedings{song2024eto,
  title={{ETO}: Trial and Error: Exploration-Based Trajectory Optimization of {LLM} Agents},
  author={Song, Yifan and Yin, Da and Yue, Xiang and Huang, Jie and Li, Sujian and Lin, Bill Yuchen},
  booktitle={Annual Meeting of the Association for Computational Linguistics (ACL)},
  year={2024},
  doi={10.18653/v1/2024.acl-long.409}
}

@inproceedings{chae2025web,
  title={Web Agents with World Models: Learning and Leveraging Environment Dynamics in Web Navigation},
  author={Chae, Hyungjoo and Kim, Namyoung and Ong, Kai Tzu-iunn and Gwak, Minju and Song, Gwanwoo and Kim, Jihoon and Kim, Sunghwan and Lee, Dongha and Yeo, Jinyoung},
  booktitle={International Conference on Learning Representations (ICLR)},
  year={2025},
  url={https://arxiv.org/abs/2410.13232}
}

@inproceedings{qiao2024wkm,
  title={Agent Planning with World Knowledge Model},
  author={Qiao, Shuofei and Fang, Runnan and Zhang, Ningyu and Zhu, Yuqi and Chen, Xiang and Deng, Shumin and Jiang, Yong and Xie, Pengjun and Huang, Fei and Chen, Huajun},
  booktitle={Advances in Neural Information Processing Systems (NeurIPS)},
  year={2024},
  url={https://arxiv.org/abs/2405.14205}
}

@article{basu2025explorer,
  title={{EXPLORER}: Exploration-guided Reasoning for Textual Reinforcement Learning},
  author={Basu, Kinjal and Murugesan, Keerthiram and Chaudhury, Subhajit and Campbell, Murray and Talamadupula, Kartik and Klinger, Tim},
  journal={arXiv preprint arXiv:2403.10692},
  year={2024},
  url={https://arxiv.org/abs/2403.10692}
}

@article{niu2024judgerank,
  title={{JudgeRank}: Leveraging Large Language Models for Reasoning-Intensive Reranking},
  author={Niu, Tong and Joty, Shafiq and Liu, Ye and Xiong, Caiming and Zhou, Yingbo and Yavuz, Semih},
  journal={arXiv preprint arXiv:2411.00142},
  year={2024},
  url={https://arxiv.org/abs/2411.00142}
}

@inproceedings{kim2024prospector,
  title={Prospector: Improving {LLM} Agents with Self-Asking and Trajectory Ranking},
  author={Kim, Byoungjip and Jang, Youngsoo and Logeswaran, Lajanugen and Kim, Geon-Hyeong and Kim, Yu Jin and Lee, Honglak and Lee, Moontae},
  booktitle={Findings of the Association for Computational Linguistics: EMNLP},
  year={2024},
  doi={10.18653/v1/2024.findings-emnlp.879}
}

@inproceedings{xiang2024retrospex,
  title={Retrospex: Language Agent Meets Offline Reinforcement Learning Critic},
  author={Xiang, Yufei and Shen, Yiqun and Zhang, Yeqin and Nguyen, Cam-Tu},
  booktitle={Conference on Empirical Methods in Natural Language Processing (EMNLP)},
  year={2024},
  doi={10.18653/v1/2024.emnlp-main.268}
}

@inproceedings{majumder2024clin,
  title={{CLIN}: A Continually Learning Language Agent for Rapid Task Adaptation and Generalization},
  author={Majumder, Bodhisattwa Prasad and Mishra, Bhavana Dalvi and Jansen, Peter and Tafjord, Oyvind and Tandon, Niket and Zhang, Li and Callison-Burch, Chris and Clark, Peter},
  booktitle={Conference on Language Modeling (COLM)},
  year={2024},
  url={https://arxiv.org/abs/2310.10134}
}

@article{yang2024cops,
  title={{CoPS}: Empowering {LLM} Agents with Provable Cross-Task Experience Sharing},
  author={Yang, Chen and Zhao, Chenyang and Gu, Quanquan and Zhou, Dongruo},
  journal={arXiv preprint arXiv:2410.16670},
  year={2024},
  url={https://arxiv.org/abs/2410.16670}
}

@inproceedings{ha2018world,
  title={Recurrent World Models Facilitate Policy Evolution},
  author={Ha, David and Schmidhuber, J{\"u}rgen},
  booktitle={Advances in Neural Information Processing Systems (NeurIPS)},
  year={2018},
  url={https://arxiv.org/abs/1809.01999}
}

@article{chen2025selfplay,
  title={Internalizing World Models via Self-Play Finetuning for Agentic {RL}},
  author={Chen, Shiqi and Zhu, Tongyao and Wang, Zian and Zhang, Jinghan and Wang, Kangrui and Gao, Siyang and Xiao, Teng and Teh, Yee Whye and He, Junxian and Li, Manling},
  journal={arXiv preprint arXiv:2510.15047},
  year={2025},
  url={https://arxiv.org/abs/2510.15047}
}

@book{kahneman2011thinking,
  title={Thinking, Fast and Slow},
  author={Kahneman, Daniel},
  year={2011},
  publisher={Farrar, Straus and Giroux},
  isbn={978-0374275631}
}

@inproceedings{liu2023agentbench,
  title={{AgentBench}: Evaluating {LLMs} as Agents},
  author={Liu, Xiao and Yu, Hao and Zhang, Hanchen and Xu, Yifan and Lei, Xuanyu and Lai, Hanyu and Gu, Yu and Ding, Hangliang and Men, Kaiwen and Yang, Kejuan and others},
  booktitle={International Conference on Learning Representations (ICLR)},
  year={2024},
  url={https://arxiv.org/abs/2308.03688}
}

@inproceedings{yu2020pcgrad,
  title={Gradient Surgery for Multi-Task Learning},
  author={Yu, Tianhe and Kumar, Saurabh and Gupta, Abhishek and Levine, Sergey and Hausman, Karol and Finn, Chelsea},
  booktitle={Advances in Neural Information Processing Systems (NeurIPS)},
  year={2020},
  url={https://arxiv.org/abs/2001.06782}
}

\clearpage
\appendix
\section{Few-Shot Fine-Tuning Details}
\label{sec:fewshot_appendix}

\begin{figure}[H]
\centering
\includegraphics[width=\columnwidth]{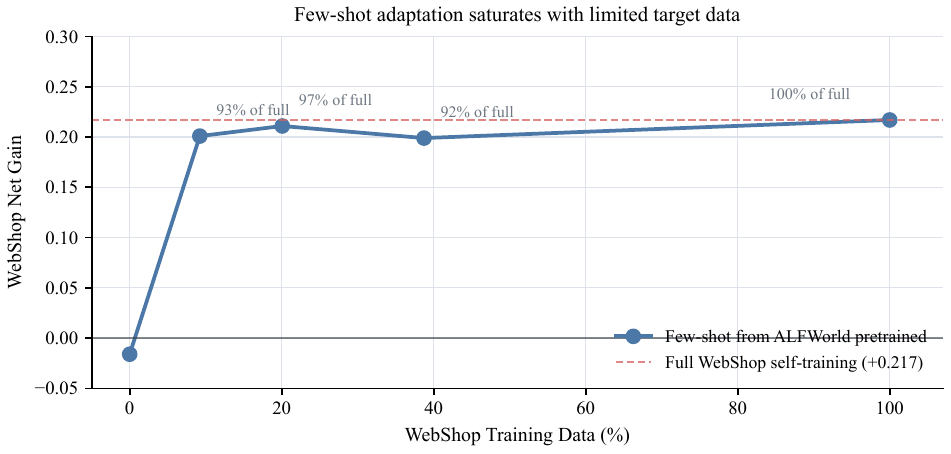}
\caption{Few-shot adaptation efficiency. An ALFWorld-pretrained DeBERTa-v3-base model
is fine-tuned on varying fractions of WebShop training data. With 9.2\% of data,
performance reaches 93\% of the full-data ceiling.}
\label{fig:fewshot}
\end{figure}

Table~\ref{tab:fewshot} and Figure~\ref{fig:fewshot} provide the full few-shot fine-tuning results
referenced in Section~\ref{sec:results_fewshot}.

\begin{table}[H]
\centering
\caption{Few-shot fine-tuning of an ALFWorld-pretrained model on WebShop data.
DeBERTa-v3-base. Training-internal evaluation. Full WebShop self-training baseline: $+0.217$.}
\label{tab:fewshot}
\small
\resizebox{\columnwidth}{!}{%
\begin{tabular}{lcc}
\toprule
\textbf{Fine-Tuning Data} & \textbf{WebShop Net Gain} & \textbf{\% of Full Performance} \\
\midrule
0\% (zero-shot)     & $-0.016$ & 0\% \\
9.2\% (965 states)  & $+0.201$ & 93\% \\
20.1\% (2,099 states) & $+0.211$ & 97\% \\
38.7\% (4,049 states) & $+0.199$ & 92\% \\
100\% (all data)    & $+0.217$ & 100\% \\
\bottomrule
\end{tabular}
}
\end{table}

\section{Per-Environment Evaluation of Two-Environment Models}
\label{sec:per_env_appendix}

Tables~\ref{tab:base_per_env} and~\ref{tab:large_per_env} report the per-environment
breakdowns for the two-environment rebalanced joint training models
(episode-level split by base episode, seed 42, unified evaluation).

\begin{table}[H]
\centering
\caption{Per-environment evaluation of DeBERTa-v3-base (184M) rebalanced joint training,
unified evaluation.}
\label{tab:base_per_env}
\small
\resizebox{\columnwidth}{!}{%
\begin{tabular}{lcccc}
\toprule
Environment & Test States & Original Top-1 & Reranked Top-1 & Net Gain \\
\midrule
ALFWorld & 1,848 & 0.107 & 0.519 & \textbf{+0.412} \\
WebShop & 2,494 & 0.124 & 0.338 & \textbf{+0.214} \\
\cmidrule{1-5}
Combined (unified eval) & 4,342 & 0.117 & 0.416 & \textbf{+0.299} \\
\bottomrule
\end{tabular}
}
\end{table}

\begin{table}[H]
\centering
\caption{Per-environment evaluation of DeBERTa-v3-large (434M) rebalanced joint training,
unified evaluation.}
\label{tab:large_per_env}
\small
\resizebox{\columnwidth}{!}{%
\begin{tabular}{lcccc}
\toprule
Environment & Test States & Original Top-1 & Reranked Top-1 & Net Gain \\
\midrule
ALFWorld & 1,848 & 0.107 & 0.435 & \textbf{+0.328} \\
WebShop & 2,494 & 0.124 & 0.112 & \textbf{$-$0.012} \\
\cmidrule{1-5}
Combined (unified eval) & 4,342 & 0.117 & 0.250 & \textbf{+0.133} \\
\bottomrule
\end{tabular}
}
\end{table}

The base model achieves an aggregate net gain of $+0.299$, while the large model
reaches $+0.133$ under the same evaluation. The large model's WebShop performance
drops to $-0.012$, confirming that with limited data, increased model capacity
does not improve cross-environment generalization.

\end{document}